\documentclass[conference]{IEEEtran}
\IEEEoverridecommandlockouts

\usepackage{cite}
\usepackage{amsmath,amssymb,amsfonts}
\usepackage{graphicx}
\usepackage{textcomp}
\usepackage{xcolor}

\usepackage{algorithm}
\usepackage{algpseudocode}
\usepackage{booktabs}  

\def\BibTeX{{\rm B\kern-.05em{\sc i\kern-.025em b}\kern-.08em
    T\kern-.1667em\lower.7ex\hbox{E}\kern-.125emX}}
\begin{document}

\title{Robust Testing for Deep Learning using Human Label Noise\\
}


\author{\IEEEauthorblockN{1\textsuperscript{st} Gordon Lim}
\IEEEauthorblockA{
\textit{University of Michigan}\\
Ann Arbor, MI, USA \\
gbtc@umich.edu}
\and
\IEEEauthorblockN{2\textsuperscript{nd} Stefan Larson}
\IEEEauthorblockA{
\textit{Vanderbilt University}\\
Nashville, Tennessee, USA \\
stefan.larson@vanderbilt.edu}
\and
\IEEEauthorblockN{3\textsuperscript{rd} Kevin Leach}
\IEEEauthorblockA{
\textit{Vanderbilt University}\\
Nashville, Tennessee, USA \\
kevin.leach@vanderbilt.edu}
}

\maketitle
\begin{abstract}
In deep learning (DL) systems, label noise in training datasets often degrades model performance, as models may learn incorrect patterns from mislabeled data. 
The area of Learning with Noisy Labels (LNL) has introduced methods to effectively train DL models in the presence of noisily-labeled datasets.
Traditionally, these methods are tested using synthetic label noise, where ground truth labels are randomly (and automatically) flipped.
However, recent findings highlight that models perform substantially worse under human label noise than synthetic label noise, indicating a need for more realistic test scenarios that reflect noise introduced due to imperfect human labeling.
This underscores the need for generating realistic noisy labels that simulate human label noise, enabling rigorous testing of deep neural networks without the need to collect new human-labeled datasets.
To address this gap, we present Cluster-Based Noise (CBN), a method for generating feature-dependent noise that simulates human-like label noise. 
Using insights from our case study of label memorization in the CIFAR-10N dataset, we design CBN to create more realistic tests for evaluating LNL methods. 
Our experiments demonstrate that current LNL methods perform worse when tested using CBN, highlighting its use as a rigorous approach to testing neural networks.
Next, we propose Soft Neighbor Label Sampling (SNLS), a method designed to handle CBN, demonstrating its improvement over existing techniques in tackling this more challenging type of noise.
\end{abstract}


\section{Introduction}
In deep learning (DL) systems, label noise in training datasets often degrades neural network (NN) performance, as NNs may learn incorrect patterns from mislabeled data~\cite{memorization2017arpit, rethinking-generalization-2021}. 
To address this challenge, the field of Learning with Noisy Labels (LNL) has introduced methods to effectively train NNs on noisy-labeled datasets.
These methods include robust loss functions~\cite{gceloss2018, sceloss2019, ma2020normalized} and sample selection strategies~\cite{yu2019does, o2unet2019, pmlr-v97-chen19g} --- a rich literature in this area exists~\cite{song2022survey}.
In general, however, these approaches leverage the fact that NNs first learn simple patterns before memorizing mislabeled examples~\cite{memorization2017arpit}.
As such, LNL methods aim to mitigate the memorization of mislabeled examples, allowing NNs to focus on learning meaningful patterns in the data.

To benchmark LNL methods in controlled settings with existing ground-truth datasets, researchers have explored ways to synthesize label noise. 
Earlier methods have applied class-dependent noise, where each class is assigned a specific probability of flipping to another class, defined by a transition matrix~\cite{confidentlearning2021}.
However, recent work with human-labeled noise from Amazon MTurk on the CIFAR-10 dataset~\cite{cifar10} revealed that models trained on human label noise, when compared again class-dependent noise with the same transition matrix, incurred reduced performance by as much as 6\%~\cite{cifar-10-100-n}.
This highlights that real-world human label noise is feature-dependent, presenting a greater challenge to NNs since they might learn the patterns of these noisy examples without needing memorization.
Consequently, there has been a shift towards evaluating on feature-dependent noise to better reflect real-world scenarios.
The polynomial margin diminishing (PMD) noise model, which generates label noise near the decision boundary of a NN trained on the original dataset, is beginning to see adoption among researchers for evaluating their LNL methods~\cite{prog_noise_iclr2021, Smart_2023_WACV, lra-diffusion-2024}.
This noise model assumes that examples along a NN's decision boundary are more ambiguous and, therefore, more likely to be mislabeled by humans.
Nonetheless, it was previously discovered that NNs and humans can have different failure modes~\cite{geirhos2018generalization, dodge2019distortions}.
In other words, what challenges NNs may not equivalently challenge humans.
Therefore, further work is needed to more closely emulate the challenges posed by \emph{human noisy labels} --- that is, data with labeling errors introduced due to imperfect labeling by human annotators --- for NNs.

In this paper, we investigate the memorization of human noisy labels on the CIFAR-10 dataset~\cite{cifar-10-100-n} to identify challenging labels that have been learned without memorization. 
Our analysis shows that certain such labels form distinct clusters in the feature space derived from CLIP~\cite{clip}, a model pre-trained on 400 million image-text pairs.
Building on these insights, we present a novel method, Cluster-Based Noise (CBN), to synthesize label noise that emulates the challenge of human noisy labels by targeting clusters within the CLIP feature space.
Specifically, CBN selects random centroids within each class's CLIP feature embeddings and flips labels within a set radius.
We show that several LNL  methods perform worse when trained on CBN compared to PMD noise at equivalent levels, highlighting CBN as a more challenging form of feature-dependent noise that can be exhibited by human annotators.
We further present a solution that improves performance on CBN by using a soft target label distribution derived from an image's nearest neighbors in the CLIP feature space. 
This demonstrates that, while our noise model presents a greater challenge, it can still be effectively managed with targeted methods.
By presenting this challenging noise model, we contribute to the literature on label noise modeling, aiding the development of LNL methods that address a broader range of noise types encountered in real-world scenarios.

\section{Preliminaries}

\subsection{Label Memorization}
\label{section:label-memorization}

The feature-dependent nature of human noisy labels suggests the presence of systematic erroneous features that NNs may learn during training without relying on memorization~\cite{cifar-10-100-n, chong-etal-2022-detecting}.
This undermines the ability of LNL methods at distinguishing between clean and noisy patterns.
To emulate this challenge posed by human noisy labels, we start by analyzing the memorization of human noisy labels in the CIFAR-10 dataset~\cite{cifar-10-100-n}.
To quantify label memorization in our study, we use the definition introduced in \cite{feldman2020memorization}:
For a learning algorithm \( \mathcal{A} \) trained on a dataset \( S = ((x_1, y_1), \ldots, (x_n, y_n)) \), the memorization of an example \( (x_i, y_i) \in S \) is defined as:

\begin{equation*}
\text{mem}(\mathcal{A}, S, i) := \Pr_{h \sim \mathcal{A}(S)}[h(x_i) = y_i] - \Pr_{h \sim \mathcal{A}(S^{\backslash i})}[h(x_i) = y_i],
\end{equation*}

where \( S^{\backslash i} \) denotes the dataset \( S \) with \( (x_i, y_i) \) removed, and the probability is computed over the randomness in the learned model \( h(\cdot) \) due to the inherent randomness of \( \mathcal{A} \), such as through random initialization.
We refer to the first term as the \textit{inclusion probability}, which is the probability that the algorithm correctly predicts the label \( y_i \) for \( x_i \) when \( (x_i, y_i) \) is part of the dataset.
Conversely, the second term, the \textit{exclusion probability}, represents the probability that the algorithm still predicts \( y_i \) for \( x_i \) when \( (x_i, y_i) \) is removed from the dataset.
If the inclusion probability is high and the exclusion probability is low for a particular example, it indicates that the algorithm heavily relies on the inclusion of that example to predict its label.
In other words, the label is memorized, as reflected by the high memorization score given by the difference between these probabilities.

Since directly estimating memorization requires retraining the NN with each training example both included and excluded, which is computationally prohibitive, we use the \textit{subsampling} estimator in \cite{feldman2020heldoutest} to approximate these probabilities.
This estimator involves training models on multiple random subsets to ensure, with high probability, that each example is included in many subsets and excluded from many others, allowing for an efficient approximation of the inclusion and exclusion probabilities.
The authors provide a theoretical bound on the estimation error using this subsampling approach, ensuring reliable approximations for memorization values~\cite{feldman2020heldoutest}.

\subsection{Feature Visualization}
\label{section:feature-visualization}

To examine the feature-level patterns of human noisy labels in CIFAR-10, we leverage CLIP to extract meaningful feature representations.
Feature extraction generally involves using a pre-trained NN to encode an image into a high-dimensional \textit{feature embedding} vector~\cite{feature-vectors-2012}.
Such feature embeddings allow us to quantitatively analyze similarities and differences across images, making it possible to uncover patterns within the data~\cite{feature-vectors-2012, algan2020label}.
We selected CLIP (Contrastive Language–Image Pretraining) as our feature extractor because it was trained on a vast dataset of 400 million image-text pairs, allowing it to capture diverse visual and semantic information to produce rich feature embeddings.
To visualize these high-dimensional feature embeddings, we use t-SNE plots~\cite{tsne}, which reduce dimensionality while preserving the relative structure and patterns within the data.

\section{Human Noisy Labels On CIFAR-10}
\label{section:case-study}

\begin{figure}[t]
    \centering
    \includegraphics[width=\linewidth]{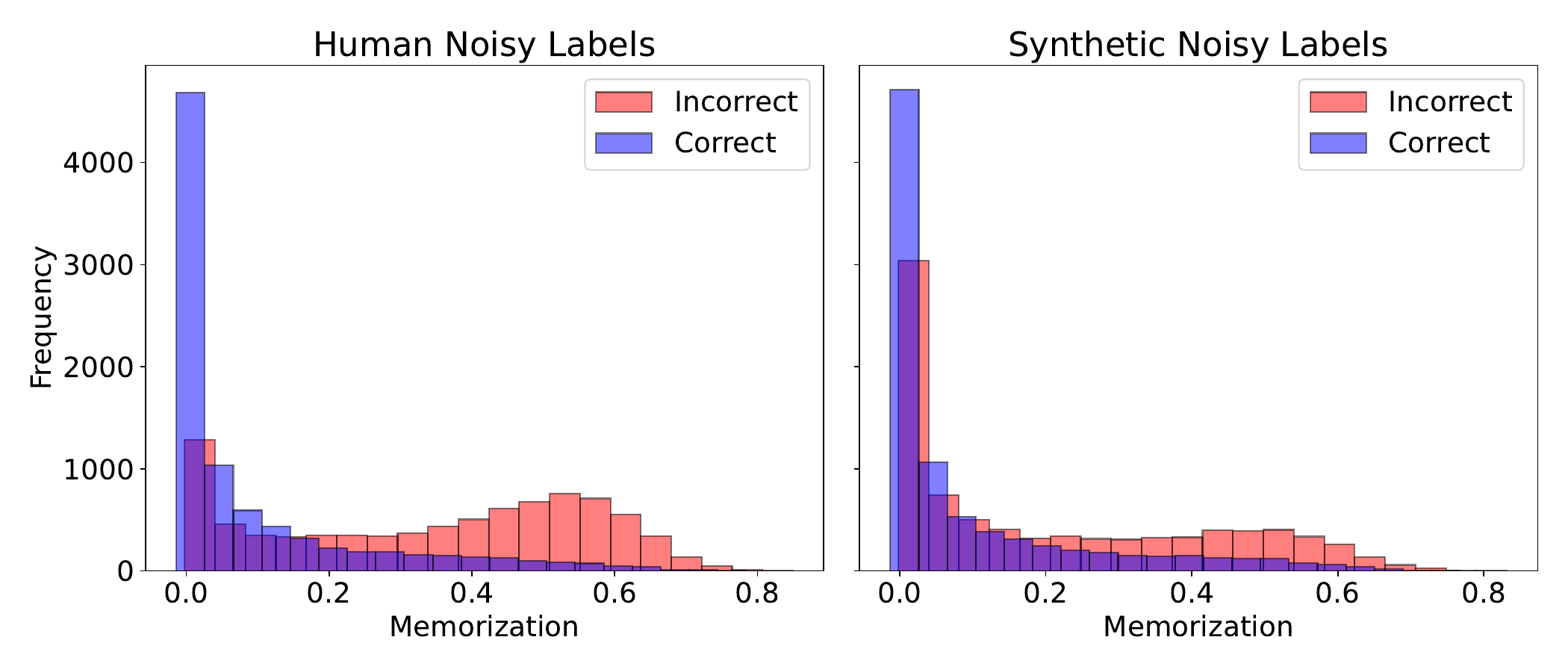}
    \caption{Memorization values for human noisy labels and synthetic class-dependent noisy labels from CIFAR-10N}
    \label{fig:memorization-histogram}
\end{figure}

In this section, we present a case study in label memorization of \emph{human noisy labels} on the CIFAR-10 dataset. 
CIFAR-10 is a widely used benchmark for image classification, consisting of 60k images at a resolution of 32x32 pixels, categorized into 10 classes: \textit{airplanes, automobiles, birds, cats, deer, dogs, frogs, horses, ships,} and \emph{trucks.} 
Each class includes 6k images, with the dataset divided into 50k images for training and 10k for testing.
CIFAR-10N~\cite{cifar-10-100-n} extends CIFAR-10 by adding human-annotated labels to the training set, collected through Amazon Mechanical Turk. 
Each training image has three human-annotated labels provided by 747 independent workers, with each worker annotating an average of 201 images.
CIFAR-10N provides five noisy-label sets by aggregating these labels in various ways, including \textit{Random} sets, where a random label is chosen per image, introducing approximately 17-18\% label noise.
Of these \textit{Random} sets, we selected the \textit{Random 1} set for our study on label memorization, as it resulted in the largest performance drop among them—up to 6\%—when trained on synthetic class-dependent noise generated with the same noise transition matrix.

We used the subsampling estimator from Feldman et al.~\cite{feldman2020heldoutest}, as described in Sec.~\ref{section:label-memorization}, to estimate memorization values~\cite{feldman2020memorization} of human noisy labels.
To further reduce computational cost, we estimated memorization values for only a subset of labels, referred to as the \textit{heldout} set.
This heldout set included both incorrect noisy labels and an equal number of correct labels for comparison.
Specifically, we trained 1,500 ResNet34~\cite{resnet} models, each with a randomly sampled 30\% of the heldout set excluded from the training data.
We repeat this procedure for the synthetic class-dependent noisy labels.
We present the histogram of memorization values in Fig.~\ref{fig:memorization-histogram}.

In both human and synthetic noisy label cases, correct labels generally exhibit lower memorization values, indicating that the model can learn these without relying heavily on memorization. 
Interestingly, incorrect labels in the synthetic noisy label set show a greater proportion with low memorization scores compared to human noisy labels. 
This observation challenges our initial intuition that human noisy labels would be more difficult due to NNs learning their erroneous patterns without needing memorization.

\begin{figure*}[htp]
    \centering

    \begin{minipage}{0.482\textwidth}
        \centering
        \includegraphics[width=\linewidth]{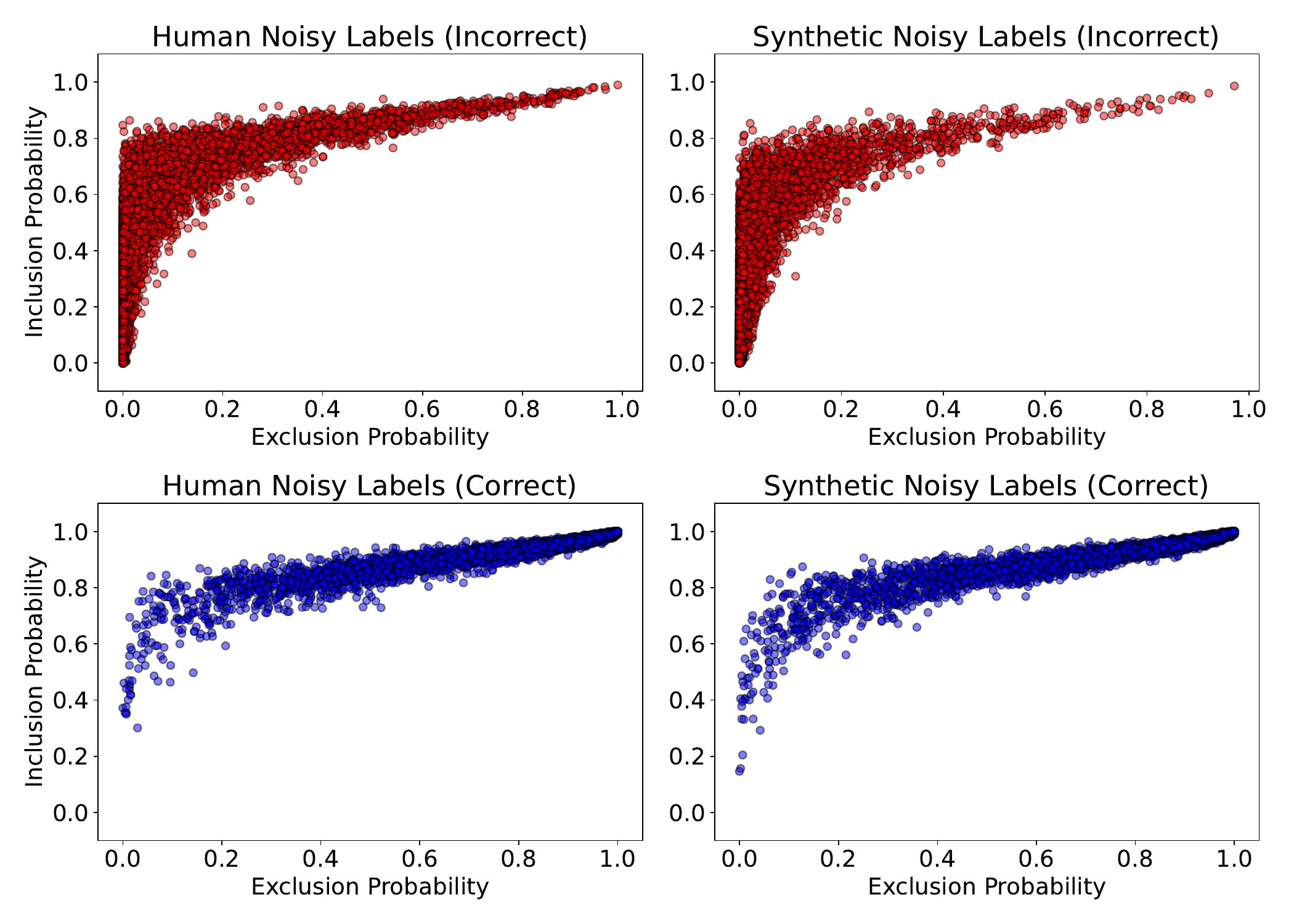}
        \caption{Scatter plot of inclusion and exclusion probabilities for human noisy labels and synthetic class-dependent noisy labels from CIFAR-10N. The distribution is visibly more dense for human noisy labels when both probabilities exceed 0.6.
        We term these examples \textit{incorrect learned human noisy labels}, representing labels that are challenging for LNL methods because they were learned without memorization despite being incorrect.} 
        \label{fig:incl-excl-scatter}
    \end{minipage}
    \hspace{0.015\textwidth} 
    \begin{minipage}{0.482\textwidth}
        \centering
        \includegraphics[width=\linewidth]{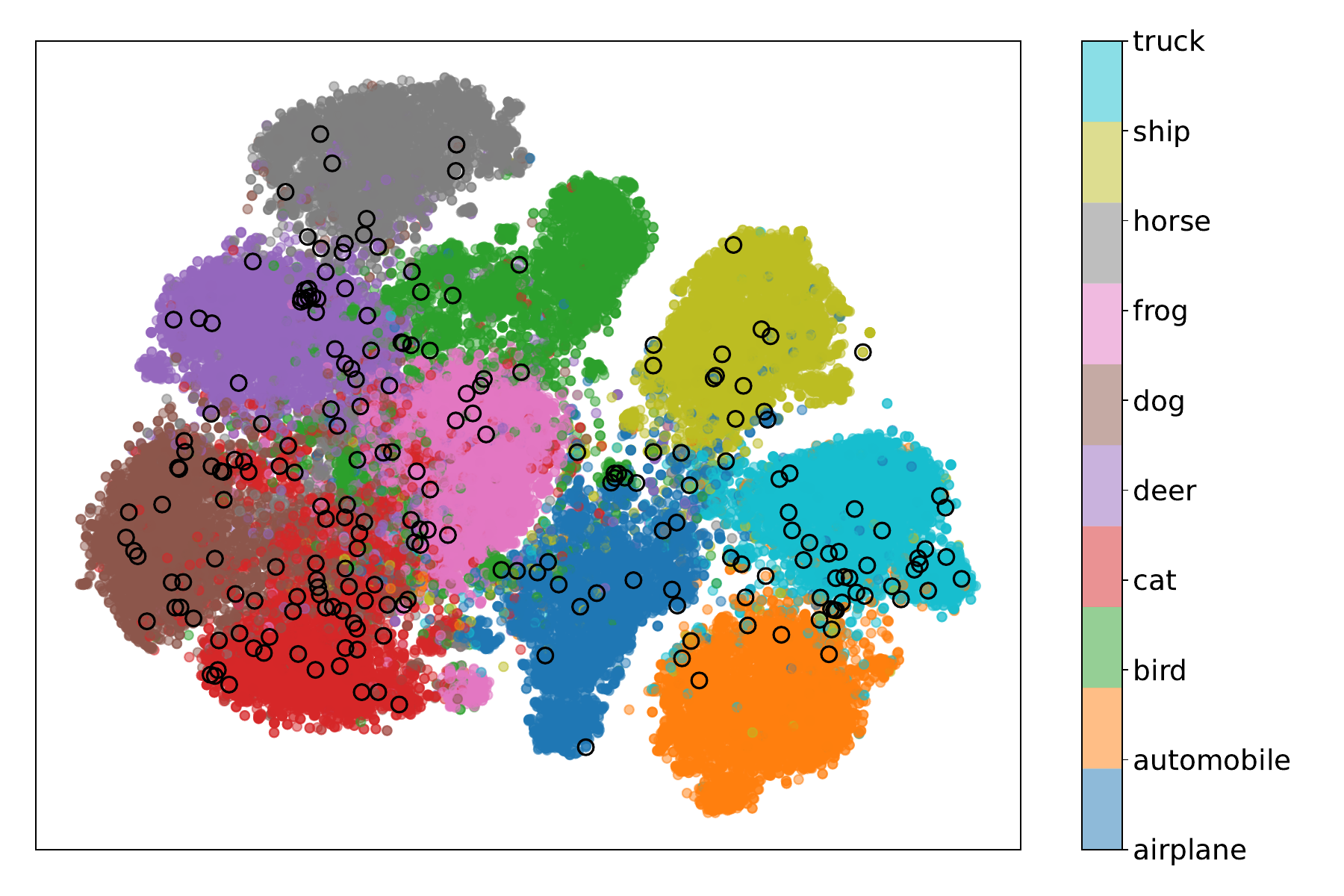}
        \caption{t-SNE plot of CIFAR-10 images' CLIP embeddings. 
        Annotated points represent \textit{incorrect learned human noisy labels.}
        There appear to be subclusters of these labels within their correct class clusters.
        } 
        \label{fig:cifar10n-clip-embeddings}
    \end{minipage}

    \vspace{1em}

    \begin{minipage}{\textwidth}
        \centering
        \includegraphics[width=\textwidth]{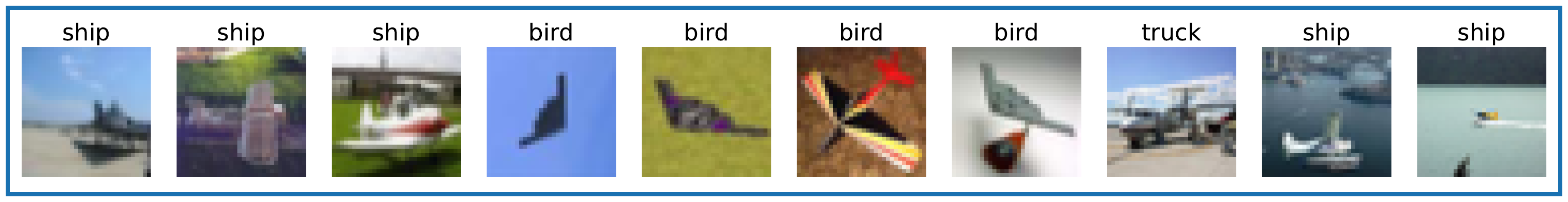}
        \vspace{0.025cm}
        \includegraphics[width=\textwidth]{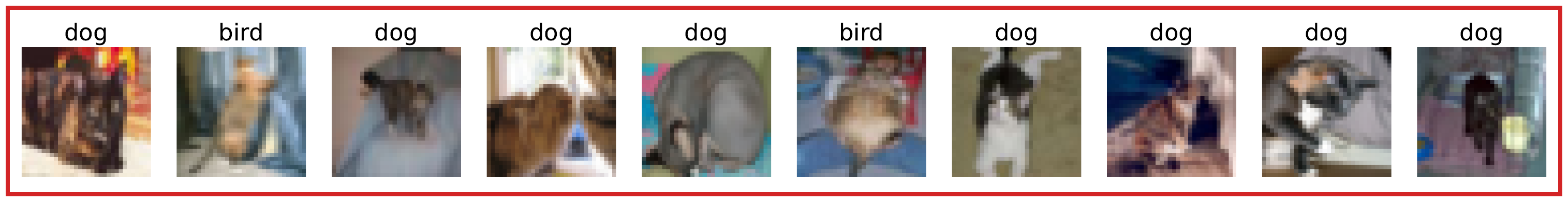}
        \vspace{0.025cm}
        \includegraphics[width=\textwidth]{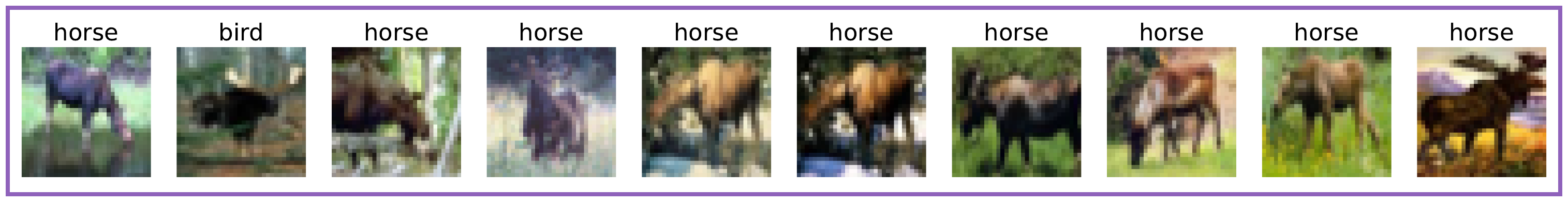}
        \vspace{0.025cm}
        \includegraphics[width=\textwidth]{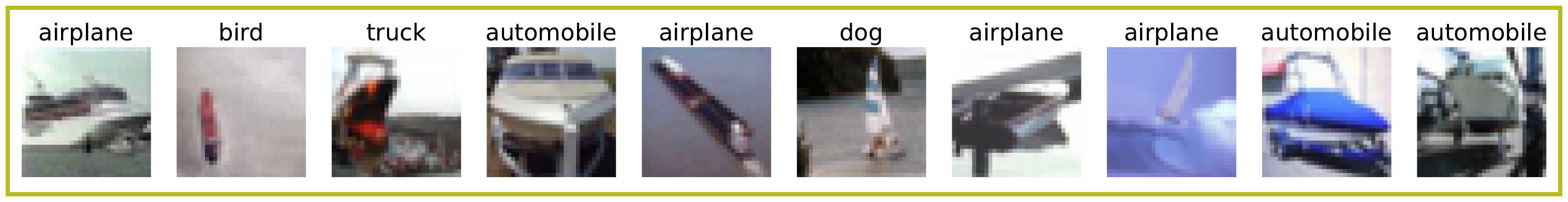}
        \vspace{0.025cm}
        \includegraphics[width=\textwidth]{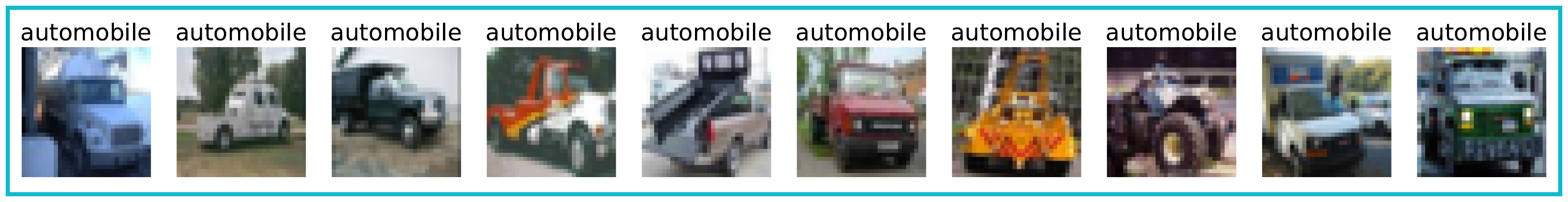}
        \caption{Top 10 closest images with \textit{incorrect learned human noisy labels} within the classes airplane (1st row), cat (2nd), deer (3rd row), ship (4th row), and truck (5th row), identified by pairwise distance in the CLIP feature space. 
        The incorrect human noisy labels are displayed above each image. 
        Bounding box colors correspond to the color coding of the given CIFAR-10 labels in Fig.~\ref{fig:cifar10n-clip-embeddings}.}
        \label{fig:cifar10n-clusters}
    \end{minipage}
\end{figure*}

To explore this further, we plot the distribution of inclusion and exclusion probabilities separately for both cases, as shown in Fig.~\ref{fig:incl-excl-scatter}.
By visualizing these probabilities separately, we uncover new insights beyond previous work~\cite{feldman2020memorization}.
Memorization, as defined in Sec.~\ref{section:label-memorization}, is calculated by the difference between inclusion and exclusion probabilities, creating two scenarios for low memorization scores.
First, when both probabilities are high (top right of the plot), the model does not rely on the example to predict its label, indicating learning without memorization—a common pattern in correct labels across both human and synthetic noisy labels.
Second, when both probabilities are low (bottom left of the plot), the model struggles to predict the label regardless of the example's inclusion, suggesting difficult or outlier examples that it fails to learn or even memorize.
Comparing incorrect labels, we observe a greater density of points in the top-right region for human noisy labels, implying that more human noisy labels are learned without memorization.
In contrast, synthetic noisy labels tend to be sparse in this region. 
This discrepancy highlights that synthetic noisy labels are often not learned by the model at all, thus posing less of a challenge for LNL methods.

To further analyze these patterns, we focus on incorrect human noisy labels with both inclusion and exclusion probabilities exceeding a threshold of 0.6—a region where human noisy labels exhibit a visibly higher density.
We term these examples \textit{incorrect learned human noisy labels}.
This set is visualized in the CLIP feature space of CIFAR-10 using a t-SNE plot in Fig.~\ref{fig:cifar10n-clip-embeddings}.
We further present the top 10 closest images with incorrect learned human noisy labels within select classes, selected by pairwise distance in Fig.~\ref{fig:cifar10n-clusters}.
These visualizations offer a couple powerful insights. 
First, we observe subclusters of these incorrect learned human noisy labels within the clusters of images belonging to the correct class, particularly prominent in the \textit{deer} and \textit{cat} categories.
Second, the human noisy label on these images tend to correspond to that of the cluster nearest to it.
For example, in the \textit{deer} cluster where a tight subcluster of incorrect learned human noisy labels exists, the human noisy label is often \textit{horse}, the nearest other cluster.
Similarly, in the \textit{airplane} cluster where there is a tight subcluster, the human noisy label is often \textit{ship} or \textit{bird}, which are the two closest other clusters.
We note that the \textit{deer} examples in Fig.~\ref{fig:cifar10n-clusters} may be mislabeled as they appear to resemble moose and thus may be out-of-distribution. 
Nonetheless, it remains interesting that their mislabeling, which led the model to learn erroneous features, can be represented as such in the CLIP feature space.
In the remainder of the paper, we will use these insights to motivate a new approach to emulate the challenge of human noisy labels.

\section{Method}

In this section, we present our novel algorithm, Cluster-Based Noise (CBN) to synthesize noisy labels that can emulate the challenge of real-world human noisy labels, that is to be able to be learned by a model without memorization as we have seen in Sec.~\ref{section:case-study}.
Then, we propose our LNL solution Soft Neighbor-Sampled Labeling in Sec.~\ref{section:snsl}, specifically developed to address this noise setting.
In Sec.~\ref{section:experiments} we benchmark our method against several LNL method and show empirical results of our method's effectiveness over existing methods.

\subsection{Cluster-based Noise}
\label{section:cbn}

\begin{algorithm}
\caption{Cluster-based noising}
\begin{algorithmic}[1]
\State \textbf{Input:} 
\State \hspace{1em} \(\mathcal{D} = \{(x_i, y_i)\}_{i=1}^n\): dataset
\State \hspace{1em} \(\mathcal{C} = \{c_i\}_{i=1}^n\): t-SNE transformed CLIP embeddings
\State \hspace{1em} \(\mathcal{Y} = \{y_i\}\): set of unique labels
\State \hspace{1em} \(n\): number of subcluster centroids
\State \hspace{1em} \(r\): radius for label flipping
\State \textbf{Output:} 
\State \hspace{1em} \(\tilde{y}\): noisy labels
\State

\State Initialize \(\tilde{y} \gets \{y_i\}_{i=1}^n\)
\\

\For{each label category \(y \in \mathcal{Y}\)}
    \State Initialize centroid as the mean of embeddings
    \indent \(u_y \gets \text{mean}(c_i \,|\, y_i = y)\) 
    \State Set \(v_{y,1}, \ldots, v_{y,n}\) as random subcluster centroids
\EndFor

\\
\For{each label category \(y \in \mathcal{Y}\)}
    \For{each data point \(x_i\) where \(y_i = y\)}
        \For{each subcluster centroid \(v_{y,j}\), \(j=1,\ldots,n\)}
            \If{distance \(d(x_i, v_{y,j}) < r\)}
                \State Set \(\tilde{y_i}\) to the label of the closest centroid \(u_{y'}\) where \(y' \in \mathcal{Y} \setminus \{y\}\)
            \EndIf
        \EndFor
    \EndFor
\EndFor

\State \textbf{return} \(\tilde{y}\)

\end{algorithmic}
\label{algorithm:cbn}
\end{algorithm}

\begin{figure*}[t]
    \centering
    \includegraphics[width=\textwidth]{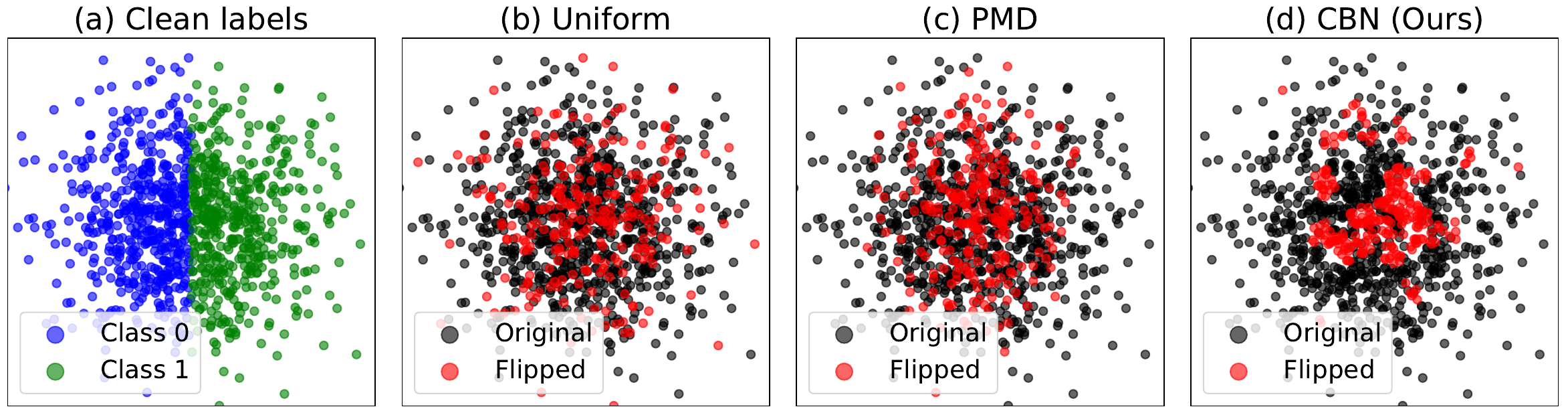}
    \caption{Comparison of noise functions at the same noise rate, visualized following~\cite{prog_noise_iclr2021}. (a) Clean labels: Gaussian blob of data labeled by a vertical decision boundary. (b) Uniform: each point has an equal probability of flipping labels. (c) PMD: points near the decision boundary have a higher probability of having its label flipped. (d) CBN (ours): labels are flipped within tight clusters of similar points.}
    \label{fig:cbn-vs-pmd}
\end{figure*}

In Sec.~\ref{section:case-study}, we found that challenging human noisy labels often form tight subclusters in their CLIP feature space.
This differs from the recently adopted PMD noise model~\cite{prog_noise_iclr2021, Smart_2023_WACV, lra-diffusion-2024}, which generates feature-dependent noise along a model's decision boundary.
Although similar challenging noise patterns appear in Fig.~\ref{fig:cifar10n-clip-embeddings}, they co-exist with the subcluster pattern.
Cluster-Based Noise(CBN) randomly selects \( n \) subcluster centroids in the t-SNE CLIP feature space, then flips labels within a specified radius \( r \) to the label of the nearest other cluster.
See our pseudocode in Algorithm~\ref{algorithm:cbn} for a detailed explanation of our algorithm. 
We compare CBN with PMD and Uniform noise using synthetic data in Fig.~\ref{fig:cbn-vs-pmd}.
Since the distribution of an unseen dataset's CLIP feature embeddings is unknown, we acknowledge a limitation: the parameters \( n \) and \( r \) must be tuned to achieve a target label noise rate.
However, we note as well that PMD noise involves a similar parameter-tuning process.

\begin{figure*}[h]
    \centering
    \includegraphics[width=\textwidth]{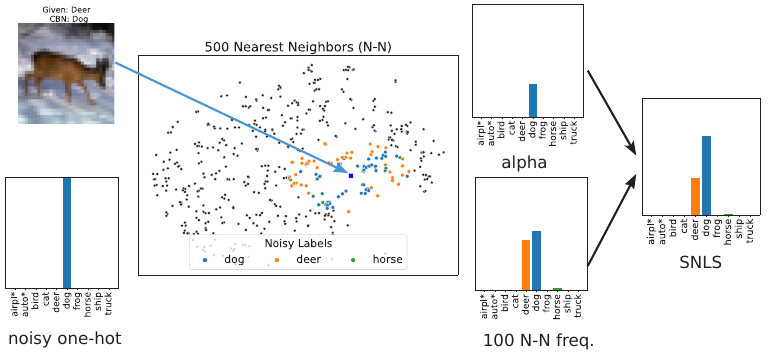}
    \caption{Illustration of our Soft Neighbor-Sampling Labeling (SNLS) technique applied to a CIFAR-10 example image. 
    The noisy one-hot label for a \textit{deer} image contains only the incorrect \textit{dog} label information. 
    SNLS generates a soft label by constructing a frequency distribution from the 100 nearest neighbors (N-N) in CLIP feature space. 
    In the scatter plot, 500 N-N are displayed, with only the closest 100 colored according to their noisy labels. 
    The final SNLS label combines this frequency distribution with an \(\alpha\) parameter representing trust in the given dataset label.
    This approach captures both the incorrect \textit{dog} and correct \textit{deer} label information, allowing the model to remain uncertain about learning the incorrect \textit{dog} label from the image.}
    \label{fig:snls-overview}
\end{figure*}

\subsection{Soft Neighbor-Sampled Labeling}
\label{section:snsl}

To address the proposed noise setting, we introduce a soft labeling technique based on label-retrieval augmentation (LRA)~\cite{lra-diffusion-2024}, which utilizes a dataset's CLIP feature embeddings instead of traditional one-hot label encoding.  
This approach assumes that neighboring embeddings, due to the design of the feature extractor (see Sec.~\ref{section:feature-visualization}), are likely to share the same label~\cite{lra-diffusion-2024}.  
In contrast to the original LRA method, which sampled a single label from \( k = 10 \) to 50 nearest neighbors, we sample from a larger neighborhood of \( k = 100 \) nearest neighbors and construct a soft label distribution incorporating information from all \( k \) neighbors.
We select \( k = 100 \) to capture richer label information from further-out neighbors, based on the assumption that in a tight cluster of incorrectly labeled examples, the further-out neighbors in the CLIP feature space may provide signals about the true label.
Additionally, we introduce an \(\alpha\) parameter representing the trust in the given label, which can be estimated by the curators of a given dataset, and combine it with the soft labeling distribution.  
We present an illustration of the technique in Fig.~\ref{fig:snls-overview}.  
The example in Fig.~\ref{fig:snls-overview} also demonstrates a particular case where our approach would excel.  
For an image of a deer that was incorrectly assigned the noisy label \textit{dog}, the one-hot noisy label provides only the incorrect label information.  
Moreover, neighboring examples also have the incorrect noisy label.  
In this situation, sampling a single label from the 10 nearest neighbors, as in~\cite{lra-diffusion-2024}, would still result in only capturing the noisy label information.  
In our approach, by sampling a larger neighborhood, we are able to capture the correct \textit{deer} signal in our final SNLS soft label.  
Thus, the model can leverage this uncertainty embedded in the soft label distribution to avoid learning features associated with dogs for this image.  
Our method can easily be used on top of any neural network architecture. 

\section{Experiments and Results}
\label{section:experiments}

We evaluate several Learning with Noisy Labels (LNL) methods on CIFAR-10 and CIFAR-100 datasets with varying noise levels and noise types.
CIFAR-100, similar to CIFAR-10, contains 100 classes instead of 10, offering a more challenging evaluation setting~\cite{cifar10}.
For both datasets, we apply label noise only to the original training dataset, while evaluation is conducted on the clean test set to accurately evaluate model performance under noisy training conditions.
We used Poly Margin Diminishing (PMD) and Class-Dependent Noise (CBN) at noise levels of 35\% and 75\%.
The methods tested include Cross Entropy (Standard), Co-teaching+~\cite{coteaching_plus}, Generalized Cross Entropy (GCE)~\cite{gceloss2018}, Progressive Label Correction (PLC)~\cite{prog_noise_iclr2021}, and LRA-Diffusion~\cite{lra-diffusion-2024}.
We use publicly available code from their respective repositories, running each method with default parameters.
Additionally, we evaluate our proposed soft labeling technique, SNLS, when used with the LRA-Diffusion architecture to compare its performance against the current state-of-the-art.
For our experiments with SNLS, we set \(\alpha = 0.30\) as a conservative lower bound estimate of clean labels in the dataset. 
To ensure reliability, each experiment is repeated three times with different random seeds, and we report both the mean and standard deviation of the results.

Our findings show that all methods exhibit lower test accuracies when trained on CBN noise compared to PMD noise.
For CIFAR-10 with a 35\% noise level, the performance drop from PMD to CBN ranges from 4.54\% to 8.96\%, and it worsens at a 70\% noise level, with a decrease from 13.90\% to 30.11\%. 
This performance gap is even more pronounced in CIFAR-100, where CBN greatly reduces accuracy.
Our results underscore that our CBN noise setting presents a more challenging scenario.

SNLS improves LRA-diffusion across all noise settings, consistently achieving the best performance. 
Although the improvement under PMD noise is modest—about a 0.19\% increase at 35\% noise for Standard—the gains for CBN are more substanstial, reaching a 1.03\% improvement at 35\% noise.
This contrast highlights the limitations of previous research, which assumed class-dependent and PMD noise and has not effectively addressed the challenges posed by CBN noise
Our findings indicate a strong need for future studies to assess methods under the CBN noise model, where current approaches still leave room for improvement.
Additionally, our results suggest that SNLS is a promising strategy to help models maintain uncertainty when learning from incorrect noisy labels in CBN noise.
\begin{table*}[h!]
    \centering
    \caption{Test accuracy (\%) of different methods across CIFAR-10 and CIFAR-100 datasets with varying noise levels and noise types}
    \begin{tabular}{lcccccccc}
        \toprule
        & \multicolumn{4}{c}{CIFAR-10} & \multicolumn{4}{c}{CIFAR-100} \\
        \cmidrule(lr){2-5} \cmidrule(lr){6-9}
        & \multicolumn{2}{c}{35\% Noise} & \multicolumn{2}{c}{70\% Noise} & \multicolumn{2}{c}{35\% Noise} & \multicolumn{2}{c}{70\% Noise} \\
        \cmidrule(lr){2-3} \cmidrule(lr){4-5} \cmidrule(lr){6-7} \cmidrule(lr){8-9}
        & PMD & CBN & PMD & CBN & PMD & CBN & PMD & CBN \\
        \midrule
        Standard         & 84.40 ± 0.18 & 75.44 ± 0.13 & 46.59 ± 0.33 & 27.22 ± 0.21 & 63.42 ± 0.15 & 46.17 ± 0.08 & 47.13 ± 0.13 & 17.48 ± 0.24 \\
        Co-teaching+~\cite{coteaching_plus}     & 67.08 ± 0.20 & 60.98 ± 0.45 & 35.35 ± 0.70 & 18.32 ± 0.14 & 55.09 ± 0.15 & 39.08 ± 0.11 & 39.36 ± 0.03 & 12.18 ± 0.09 \\
        GCE~\cite{gceloss2018}              & 84.70 ± 0.10 & 77.73 ± 0.28 & 39.06 ± 0.66 & 25.16 ± 0.45 & 63.08 ± 0.25 & 39.60 ± 0.56 & 43.00 ± 0.25 & 12.59 ± 0.41 \\
        PLC~\cite{prog_noise_iclr2021}              & 86.11 ± 0.02 & 80.51 ± 0.19 & 42.66 ± 2.08 & 23.06 ± 4.08 & 62.23 ± 0.17 & 42.67 ± 0.15 & 47.86 ± 0.24 & 12.69 ± 0.37 \\
        LRA-Diffusion~\cite{lra-diffusion-2024}    & 97.12 ± 0.10 & 91.74 ± 0.48 & 47.17 ± 2.00 & 18.60 ± 1.29 & 77.86 ± 0.43 & 50.34 ± 0.34 & 57.18 ± 0.81 & 11.76 ± 0.24 \\
        \textbf{LRA-Diffusion+SNLS}   & \textbf{97.31 ± 0.03} & \textbf{92.77 ± 0.18} & \textbf{49.16 ± 2.01} & \textbf{19.05 ± 0.49} & \textbf{78.89 ± 0.28} & \textbf{58.80 ± 0.51} & \textbf{62.41 ± 0.51} & \textbf{15.13 ± 0.20} \\
        \bottomrule
        \vspace{0.1cm}
    \end{tabular}
\end{table*}

\section{Related Work}

In this section, we review work related to the methods introduced in our paper.
First, we examine other types of label noise benchmarks.
Then, we discuss previous approaches that use a soft label distribution and highlight what makes ours unique.

\noindent \textbf{{Label Noising.}} 
It has been previously explored that naïve methods for synthesizing label noise in benchmarking Label Noise Learning (LNL) methods---such as adding random noise or class-dependent noise---are insufficient to capture the complexities of real-world human labeling errors, which can be feature-dependent~\cite{cifar-10-100-n, chong-etal-2022-detecting}.
Consequently, research has focused on developing better noisy label sets to guide the development of LNL methods that are robust against real-world noise conditions.
One approach is to directly collect human labels~\cite{song2019selfie}, but this does not allow for controlled evaluation.
To address this limitation, collecting multiple human labels enables sampling of label errors until a desired noise level is achieved~\cite{cifar-10-100-n, chong-etal-2022-detecting}.
However, this process is costly and leads to a limited availability of image classification datasets with multiple human annotations, especially for specialized domains beyond animals and vehicles.
Recent research explored generating feature-dependent noise by using the features learned by a model to flipping labels to similar classes based on the model's class probabilities~\cite{algan2020label, prog_noise_iclr2021}.
Perhaps most similar to our work is locally concentrated noise (LLN)~\cite{inouye2017hyperparameter}.
The original LLN study focused on synthetic and tabular data, testing traditional machine learning methods like KNN, SVM, and decision trees.
It was extended in~\cite{algan2020label} to incorporate a learned student network within the knowledge distillation framework~\cite{hinton2015distilling}.
However, our work leverages the more powerful CLIP model, which can better represent similar features.
Motivated by a real-world case study on memorization, we also flip labels to the closest class, unlike their method that uniformly samples corrupted labels.
Furthermore, their approach involves only one local subcluster, whereas our method is more challenging due to the presence of multiple subclusters.

\noindent \textbf{{Soft labeling.}}
Traditionally, neural networks are trained using one-hot encoded labels, where each example places all of its probability on its given label.  
Early forms of soft labeling, such as label smoothing, redistributed some probability from the given label to other categories , helping to reduce overconfidence by penalizing overfitting to single hard labels~\cite{szegedy2016rethinking}.  
To build a feature-aware soft label distribution, MixUp~\cite{zhang2018mixup} and CutMix~\cite{yun2019cutmix} combined pairs of images with known weights, then using these weights to build a soft label distribution. 
Other approaches use deep learning to learn a soft label distribution which would optimizes training~\cite{hinton2015distilling, meta-soft-label}. 
Another method involved using a crowd of annotators, where their vote distribution for each label was normalized into a soft label distribution~\cite{nguyen2014learning, cifar10n}.  
This approach captured human uncertainty more effectively, with distributions reflecting the varying features of the images.  
Our approach to soft labeling builds on these past works by leveraging the representation of challenging human label errors in the pretrained CLIP feature space, thereby eliminating the need to train a new model that might inherit dataset biases or to collect explicit human uncertainty labels.

\section{Conclusion}
This paper presents the first study to examine the memorization values of human noisy labels, introducing a new perspective to analyzing these values by distinguishing between inclusive and exclusive probabilities. 
This approach allowed us to visualize incorrect human noisy labels that are \textit{learned} by the model---labels that, despite being erroneous, behave like clean labels and are particularly challenging for learning with noisy labels (LNL). 
We observe in their CLIP feature space that such challenging labels form within subclusters of their respective class clusters. 
Motivated by these findings, we introduce cluster-based noise (CBN) using t-SNE of CLIP embeddings as a new benchmark for evaluating LNL robustness. 
Our experimental results indicate that several existing LNL methods perform worse on CBN than on poly margin diminishing (PMD) noise, which assumes label noise primarily at decision boundaries. 
To address this, we propose SNSL, a method that creates a soft label distribution based on the 100 nearest neighbors in the CLIP embedding space, showing improved performance on CBN. 
However, further improvements are needed, and we recommend that future LNL research consider CBN as an evaluation metric to better develop LNL methods that can withstand various types of feature-dependent label noise, as encountered in real-world conditions.

\bibliographystyle{IEEEtran}
\bibliography{ieee}

\end{document}